\documentclass{article}
\title{Bibliography management: BibTeX}
\author{Overleaf}

\usepackage{PRIMEarxiv}

\usepackage[utf8]{inputenc} 
\usepackage[T1]{fontenc}    
\usepackage{hyperref}       
\usepackage{url}            
\usepackage{booktabs}       
\usepackage{amsfonts}       
\usepackage{nicefrac}       
\usepackage{microtype}      
\usepackage{lipsum}
\usepackage{fancyhdr} 
\usepackage{booktabs}
\usepackage{graphicx}
\usepackage[table,xcdraw]{xcolor}
\graphicspath{{media/}}     

\title{Evaluating CNN with Oscillatory Activation Function
}

\author{
  Jeevanshi Sharma \\
  Department of Electrical Engineering \\
  Aligarh Muslim University, Aligarh\\
  \texttt{sharma.jivanshi@gmail.com} \\
}

\begin{document}
\maketitle

\begin{abstract}
The reason behind CNN's capability to learn high-dimensional complex features from the images is the non-linearity introduced by the activation function. Several advanced activation functions have been discovered to improve the training process of neural networks, as choosing an activation function is a crucial step in the modeling. Recent research has proposed using an oscillating activation function to solve classification problems inspired by the human brain cortex. \cite{r41} This paper explores the performance of one of the CNN architecture ALexNet on MNIST and CIFAR10 dataset using oscillatory activation function (GCU) and some other commonly used activation functions like ReLu, PReLu, Mish. 
\end{abstract}

\keywords{Activation Function \and Convolutional Neural Networks \and Growing Cosine Unit \and XOR Function}

\section{Introduction}
The convolutional neural network is a type of feed-forward multi-layer neural network, which is generally used for image datasets. It is composed of a convolution layer (Conv), a down-sampling layer (or pooling layer), and a full-connection layer (FC). The convolution layer is responsible for the extraction of the different features from the input images, which gets convoluted by several filters on the convolution layer. Then the feature is blurred by the down-sampling layer. Finally, a set of eigenvectors is acquired through a full connection layer. Activation functions are an important part of CNN's architecture and its performance. CNN's have been successfully in practically various fields \cite{r1} \cite{r2} \cite{r3} \cite{r4}. The general architecture of a convolutional neural network is represented in Figure 1.

\begin{figure}[htbp]
\centerline{\includegraphics[scale = 0.22]{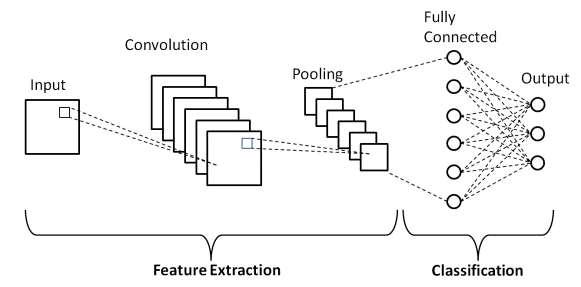}}
\caption{CNN Architecture}
\label{fig1}
\end{figure}

The practical implementation of a CNN model still demands improvement. Many studies have been done on different image classification methods\cite{r5} \cite{r6}, design of adaptive learning rate \cite{r7}, design of drop-out layer \cite{r8} \cite{r9}, and at the core research on better activation functions for mapping data better \cite{r10} \cite{r11} \cite{r12}. Neural networks are non-linear.  It is their non-linearity that makes them powerful.  As natural images have highly co-related pixel values, this attribute can only be better exploited with the use of non-linear functions. Linear activation functions make no use of the multi-layer CNNs. \cite{r13}

Over the years, there has been a lot of significant development in CNN architectures to mitigate problems regarding computational efficiency, error rates, and other improvements. Architectures like LeNet-5 \cite{r14}, AlexNet \cite{r15}, Inception \cite{r16}, VCGNet \cite{r17}, ResNet \cite{r18}, etc. were put forward to deal with issues of overfitting, issues with high-resolution images, etc. Past studies have shown to achieve better results when these architectures are optimized with different activation functions \cite{r19} and optimization algorithms. Along with this, many variants of activation functions like ReLu and different combinations of activation functions have been proposed to achieve better results with these architectures. \cite{r20} \cite{r21} \cite{r22}. In particular, the use of the ReLU like activation functions can result in faster training compared to saturating sigmoidal type activation functions because these activation functions do not saturate for a wider range of inputs and avoid the vanishing gradient problem. The classification of activation functions is primarily based on whether they can modify their shape during training. \cite{r23} Oscillatory activation functions which are inspired by biological comprises of multiple hyperplanes in their decision boundaries, which enables neurons to make more complex decisions. In \cite{r24}, the author suggests that deep neural networks with oscillatory activation functions might partially bridge the gap present in biological and artificial neural networks. One such function is GCU (Growing Cosine Unit). \cite{r25} In contrast to the common perception of a single plane decision boundary, the GCU has infinite
evenly spaced solutions resulting in infinite evenly spaced hyperplanes, which solves the XOR problem in neural networks.

This study talks about the development of trainable activation function and their performance on CNN architectures, AlexNet and ResNet.

\section{Activation Functions}
In a neural network, neurons map the weighted sum of their input to a single output with the help of Activation Functions. The main purpose served by the activation function is to introduce non-linearity in the network. The important properties commonly shared by activation functions are non-linearity, differentiability, continuous, bounded, and zero-centering. \cite{r26} Non-linear functions are universal approximates and using a linear activation function in a multi-layer network is the same as using a single-layer network. \cite{r27} Continuously differentiable activation functions are desired when using gradient-based optimization algorithms. Activation functions are preferred to be centered at zero so that the gradients do not shift in a particular direction for gradients to be
unbiased the output of the activation function must be symmetrical at zero so that the gradients do not shift in a particular direction.  

\subsection{ReLU}

ReLU \cite{r28} was primarily used to overcome the vanishing gradient problem. ReLU is the most common activation
function used for classification problems.


\begin{figure}[htbp]
\centerline{\includegraphics[scale = 0.7]{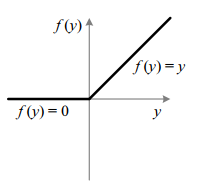}}
\caption{ReLU}
\label{fig2}
\end{figure}

\subsection{PReLU}

The disadvantage of ReLU is that it cannot adapt to sudden changes and zero the negative inputs (dying ReLU problem). PReLU generalises the traditional with a slope for negative inputs using a learnable parameter that leads to adaptive constants which can increase accuracy. \cite{r29}

\begin{figure}[htbp]
\centerline{\includegraphics[scale = 0.7]{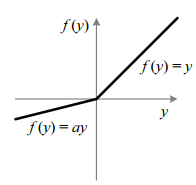}}
\caption{PReLU}
\label{fig3}
\end{figure}

\subsection{Mish}

Mish is an unbounded in positive direction and a bounded negative domain. Mish has properties like non-monotonicity and smooth profile. \cite{r30}

\begin{figure}[htbp]
\centerline{\includegraphics[scale = 0.4]{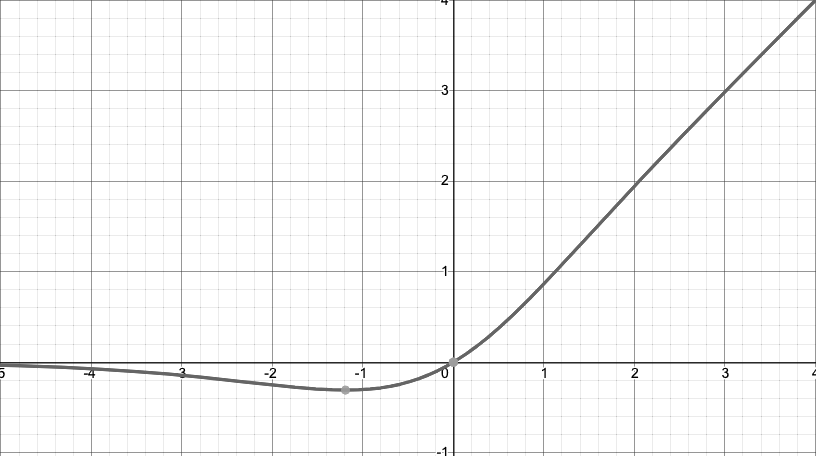}}
\caption{Mish}
\label{fig4}
\end{figure}

\subsection{GCU}

Growing Cosine Unit was introduced in 2021 \cite{r40}. It is an oscillatory activation function whose amplitude increases with increasing values. GCU has infinite evenly spaced solutions resulting in infinite evenly spaced hyper-planes, instead of a single plane decision boundary which resolves the long-rooted XOR problem as for separating classes in XOR datasets two hyper-planes are needed which required the use activation function with multiple zeros.\cite{r42}

\begin{figure}[htbp]
\centerline{\includegraphics[scale = 0.5]{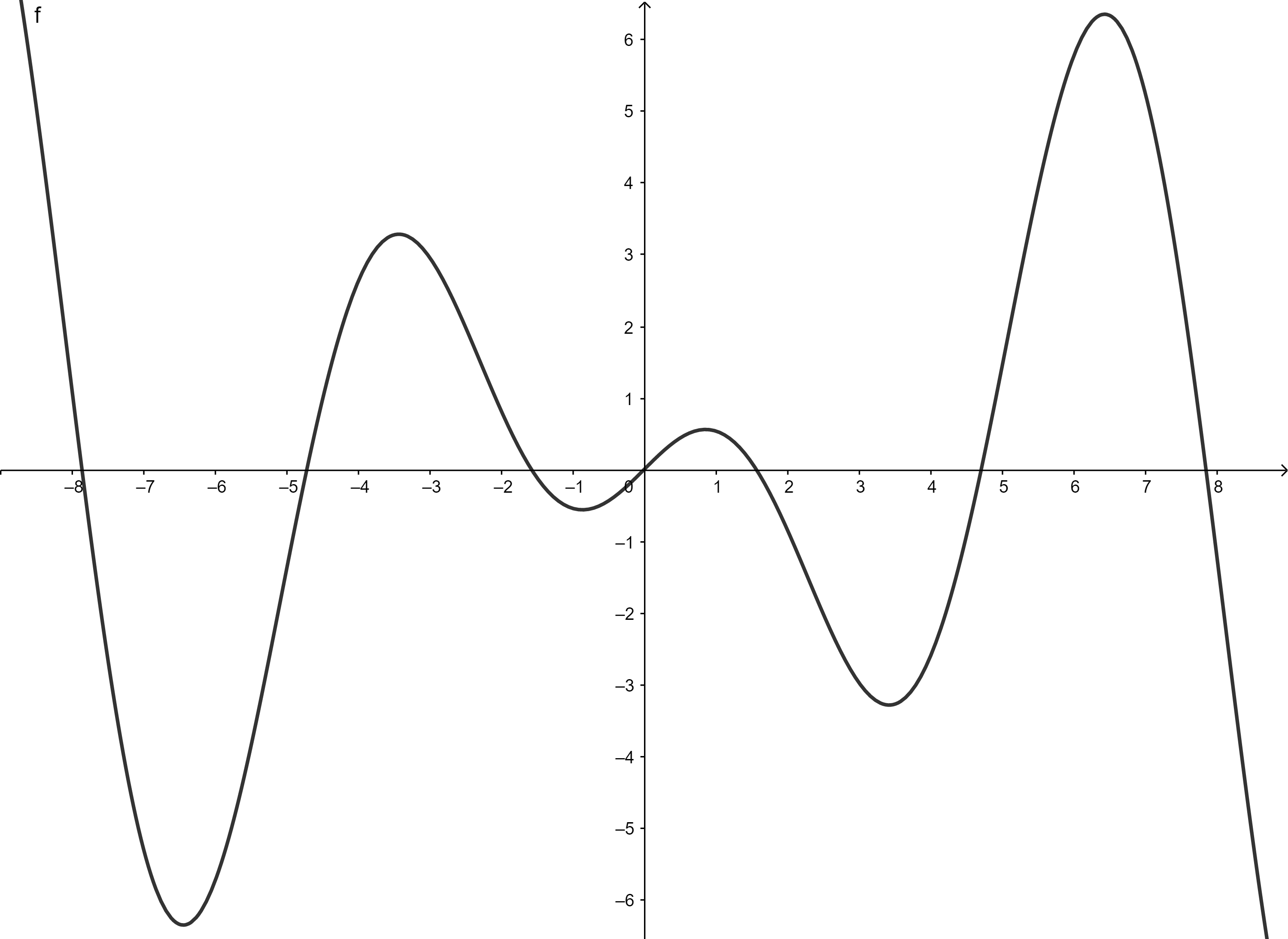}}
\caption{GCU}
\label{fig5}
\end{figure}

\section{Dataset and Methodology}

\subsection{Experimental Setup}

In this following, a comparison of the AlexNet on CIFAR10 and MNIST with GCU and other activation functions is presented. The SGD  and Adam optimizers are used on CIFAR and MNIST datasets respectively with sparse categorical crossentropy loss function (softmax classification head). For both the datasets, we experiment with the choice of activation functions at two locations in AlexNet:
one for all of the convolutional layer and the other for the dense layers.

The results for CIFAR 10 are reported on 50 epchos and 40 epochs for MNIST. 

\section{Results}

The performance of GCU with various activation functions are visualized and presented in Table 1 and Table 2 for comparing the oscillating activation functions with other prominently used non-linear activation fucntions (Mish, Prelu).

\begin{table}[h]
\centering
\resizebox{\textwidth}{!}{%
\begin{tabular}{@{}lllllll@{}}
\toprule
{\color[HTML]{000000} \textbf{Architecture}} & {\color[HTML]{000000} \textbf{Convolutional Layer}} & {\color[HTML]{000000} \textbf{Activation Dense Layer}} & \textbf{Validation Accuracy} & \textbf{Test Accuracy} & \textbf{Val Loss} & \textbf{Loss} \\ \midrule
{\color[HTML]{000000} AlexNet}               & {\color[HTML]{000000} ReLU}                         & {\color[HTML]{000000} ReLU}                            & 0.9940                       & 0.9974                 & 0.0850            & 0.0145       \\
{\color[HTML]{000000} AlexNet}               & {\color[HTML]{000000} GCU}                          & {\color[HTML]{000000} ReLU}                            & 0.9830                       & 0.9791                 & 0.0791            & 0.9830        \\
{\color[HTML]{000000} AlexNet}               & {\color[HTML]{000000} PReLU}                        & {\color[HTML]{000000} ReLU}                            &0.9945                              &0.9904                         &0.0469                   &0.0435              \\
{\color[HTML]{000000} AlexNet}               & {\color[HTML]{000000} Mish}                         & {\color[HTML]{000000} ReLU}                            &0.9940                              &0.9928                         &0.1190                   &0.0625                \\ \bottomrule
\end{tabular}%
}
\newline
\caption{Performance on MNIST}
\label{tab:mnist}
\end{table}

\begin{table}[h]
\centering
\resizebox{\textwidth}{!}{%
\begin{tabular}{@{}lllllll@{}}
\toprule
{\color[HTML]{000000} \textbf{Architecture}} & {\color[HTML]{000000} \textbf{Convolutional Layer}} & {\color[HTML]{000000} \textbf{Activation Dense Layer}} & \textbf{Validation Accuracy} & \textbf{Test Accuracy} & \textbf{Val Loss} & \textbf{Loss} \\ \midrule
{\color[HTML]{000000} AlexNet}               & {\color[HTML]{000000} ReLU}                         & {\color[HTML]{000000} ReLU}                            & 0.7011                       & 0.9777                 & 1.6989            & 0.0666        \\
{\color[HTML]{000000} AlexNet}               & {\color[HTML]{000000} GCU}                          & {\color[HTML]{000000} ReLU}                            & 0.6036                       & 0.8256                 & 1.3953            & 0.4913        \\
{\color[HTML]{000000} AlexNet}               & {\color[HTML]{000000} PReLU}                        & {\color[HTML]{000000} ReLU}                            & 0.7015                       & 0.9765                 & 1.6207            & 0.0722        \\
{\color[HTML]{000000} AlexNet}               & {\color[HTML]{000000} Mish}                         & {\color[HTML]{000000} ReLU}                            & 0.7143                       & 0.9802                 & 1.6899            & 0.0591       

\\ \bottomrule
\end{tabular}%
}
\newline
\caption{Performance on CIFAR 10}
\label{tab:cifar10}
\end{table}

\section{Conclusion}

The activation function is an important part of a Convolutional Neural Network. Many research have proposed multiple methods which can improve the performance of CNNs. From the view of activation function, this paper explores and experiment the use of oscillatory activation function (GCU) in AlexNet, and compares it with PReLu and Mish. As Mish is more computationally heavy than ReLU, we have only placed these function in the convolution layers and ReLU for the deep connected layers. GCU have shown comparable performance on both MNIST and CIFAR 10 dataset. 

\bibliographystyle{unsrt}  
\bibliography{references}

\end{document}